\documentclass{ifacconf}

\usepackage{graphicx,subfigure}  
\usepackage{natbib}        
\begin{document}
\begin{frontmatter}

\title{Sim2Real Grasp Pose Estimation for Adaptive Robotic Applications} 

\author[First,Second]{Dániel Horváth} 
\author[First]{Kristóf Bocsi} 
\author[First,Third]{Gábor Erdős}
\author[Second]{Zoltán Istenes}

\address[First]{Centre of Excellence in Production Informatics and Control, Institute for Computer Science and Control, Eötvös Loránd Research Network, Budapest, Hungary (e-mail: daniel.horvath@sztaki.hu)}
\address[Second]{CoLocation Center for Academic and Industrial Cooperation, Eötvös Loránd University, Budapest, Hungary}
\address[Third]{Department of Manufacturing
Science and Engineering, Budapest University of Technology and
Economics, Budapest, Hungary}

\begin{abstract}                
Adaptive robotics plays an essential role in achieving truly co-creative cyber physical systems. In robotic manipulation tasks, one of the biggest challenges is to estimate the pose of given workpieces. Even though the recent deep-learning-based models show promising results, they require an immense dataset for training. In this paper, two vision-based, multi-object grasp pose estimation models (MOGPE), the MOGPE Real-Time and the MOGPE High-Precision are proposed. Furthermore, a sim2real method based on domain randomization to diminish the reality gap and overcome the data shortage. Our methods yielded an 80\% and a 96.67\% success rate in a real-world robotic pick-and-place experiment, with the MOGPE Real-Time and the MOGPE High-Precision model respectively. Our framework provides an industrial tool for fast data generation and model training and requires minimal domain-specific data.
\end{abstract}


\end{frontmatter}

\begin{figure*}[hbt!]
\centering
\includegraphics[width=0.9\textwidth]{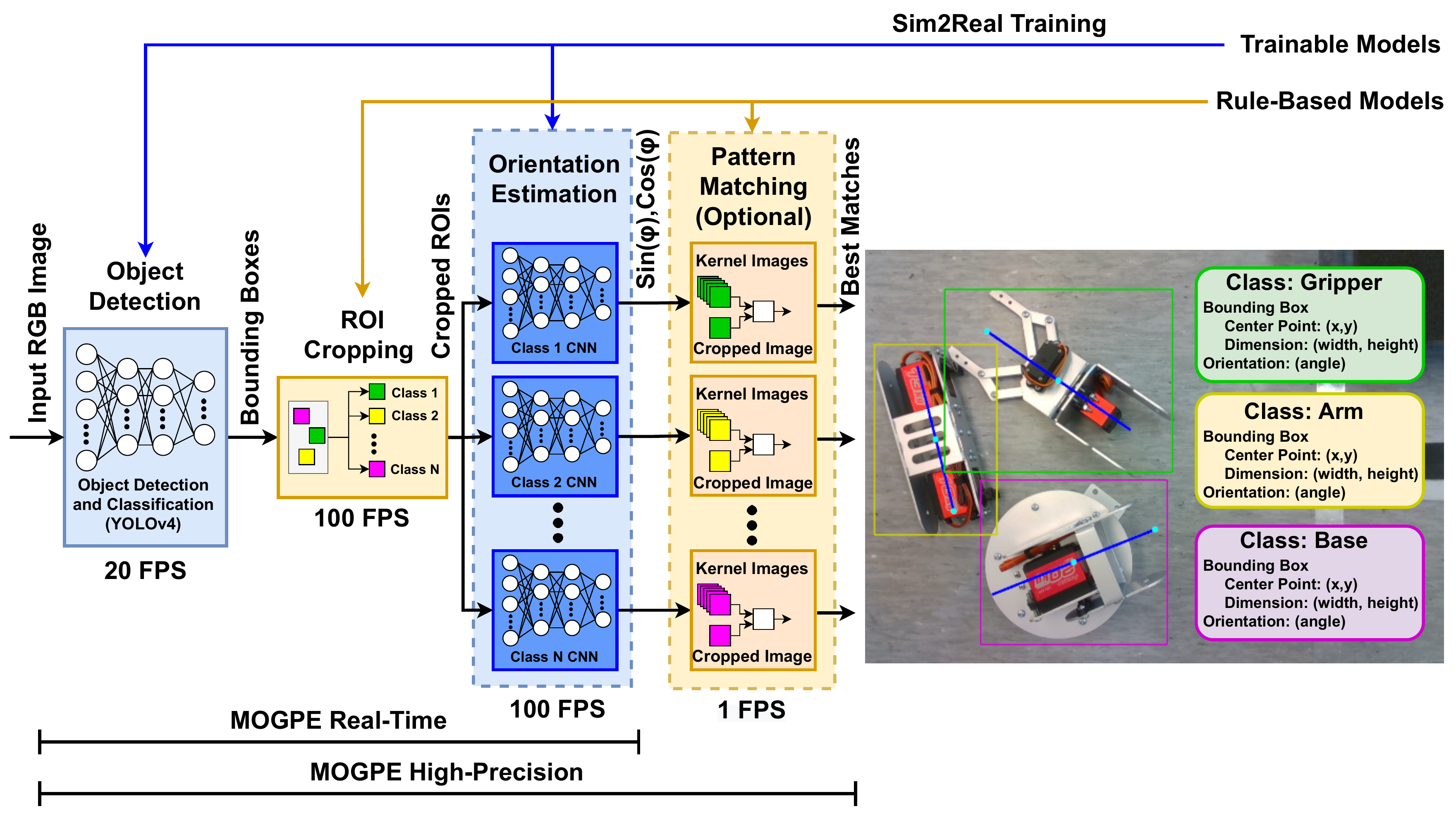}
\hfil
\caption{Illustration of our multi-object grasp pose estimation method.}
\label{fig_cover}
\end{figure*}

\section{Introduction}
Adaptive robotics aims to solve challenges arising from the concept of co-creative cyber physical systems. Traditional robotic applications can move objects or assemble parts fast and reliably in a fully-controlled environment that is well-suited for mass production. Applying these traditional applications is not economically feasible for lower volume production such as manufacturing customized products. Additionally, in many situations, robots need to work either physically alongside human workers (human-robot collaboration) or in a workflow where their input is significantly influenced by the dexterity of human workers~\citep{tian_developing_2021,zhouHumanCyberPhysical2019}.

Adaptive robots need to sense
and interpret their environment and make informed and
automatic decisions on how they maximize their targets.
Similarly to humans, vision is an efficient way to perceive the environment.

Even though deep-learning-based models revolutionized the field of computer vision, their applications in the field of robotics have obstacles. Collecting the datasets for these notoriously data-hungry learning-based models, on many occasions, is not feasible in the industry. \cite{levineLearningHandEyeCoordination2016} recorded over 1.7 million robotic grasp attempts over several months using between 6 and 14 robots at the same time.

Transfer learning\citep{weissSurveyTransferLearning2016} in the case of supervised learning can be described as creating synthetic data to train machine learning models. Thus, the tedious work of collecting and labeling data can be omitted. The model trained on synthetic data, ceteris paribus, will not generalize well for the real domain. This is called the reality gap that transfer learning aims to diminish. These methods can generally be grouped into the field of domain randomization and domain adaptation. The former methods try to diminish the reality gap by introducing artificial noise and randomness, thus the model will not overfit on the domain-specific characteristics but learn the underlying data representation of the objects. Whereas the latter approaches attempt to transform the source domain to the target domain (generating photo-realistic images as an example) or transfer the source and the target domain to a third domain.

Robotic grasping is an unsolved problem and a critical challenge of adaptive robotics in which the model not only needs to identify and locate the different parts but estimate its orientation to compute a viable grasp position. The contributions of the paper are as follows: 
\begin{itemize}
\item The proposed multi-object grasp pose estimation methods (MOGPE), the MOGPE Real-Time and MOGPE High-Precision models.
\item The synthetic data generation process with sim2real domain randomization for grasp pose estimation.
\item Our freely available implementation of the grasping pose estimation\footnote{https://git.sztaki.hu/emi/grasping-pose-estimation} and the robot control framework\footnote{https://git.sztaki.hu/emi/robot\_control\_framework}.
\end{itemize}
Our results:
\begin{itemize}
\item In our case study, the object detection model yielded a 98.78\% mAP\textsubscript{50} score, while the orientation estimation models achieved a 97.04\% success rate on average.
\item The MOGPE Real-Time (RT) model runs in real time. The object detection stage works at 20 FPS while the orientation estimation stage runs at 100 FPS.
\item In a real-world experiment of robotic grasping, the MOGPE RT model achieved an 80\% while the MOGPE High-Precision (HP) model accomplished a 96.67\% success rate. These results serve as a proof-of-concept of our approach.
\end{itemize}

This work is a continuation of our previous work~\citep{horvath_object_2022} where a sim2real framework for object detection was proposed.

\section{Problem Statement}\label{sec:problem_statement}

In this section, the problem is briefly presented alongside our approach. For a complete overview of the field the reader is refered to survey articles such as~\cite{kleeberger_survey_2020}

The problem defined as a 3.5 DoF ($x$,$y$,$\theta$,$c$) pick-and-place robot manipulation task. The planar coordinates ($x$,$y$), the $\theta$ angle of the orientation and the classes ($c$) of the objects need to be estimated. Further characteristics of the problem are as follows. The position of the plane where the objects are placed must be known. The objects are recognized only from one of their stable position. The parts are separated and all object classes are present at the training.

The given model needs to identify and locate all the different workpieces and then estimate the orientations of them. Additional challenges arise from the following circumstances. The environment is not controlled (no special illumination), and the background is not simplified (no monochromatic background). The model has access to only one RGB image, thus the 3D reconstruction of the scene is not possible. The grasp must be performed with a two-finger gripper and every object has only one grasp position. 

Our solution is a two-stage, data-driven (supervised learning), still 3D model-based method. As our aim is industrial usability, the assumption is that the availability of real-world data is limited. The majority of the training dataset is synthetic, generated by our sim2real domain randomization framework.  

\section{Related works}
The related works focus mostly on two aspects of the robotic grasping challenge:
\begin{enumerate}
    \setlength\itemsep{0em}
    \item What is the optimal model to solve the problem?
    \item How to generate training data and then transfer the knowledge to the real world?
\end{enumerate}

\cite{mahler_learning_2019} introduced Dex-Net 4.0. They use a simulator to create a training dataset for their Grasp Quality Convolutional Neural Network. Even though this approach is relatively strong in bin-picking tasks, it is less optimal for pick-and-place operations with predefined grasping positions.

\cite{tobinDomainRandomizationGenerative2018} propose an autoregressive grasp planning that gives a probability distribution over possible grasps. They used the YCB~\citep{calli_ycb_2015} dataset and in a real-world scenario, they achieved an 80\% success rate.

\cite{pashevichLearningAugmentSynthetic2019} trained a model to learn manipulation policies in a simulation using depth images and sim2real transfer. They achieved $1.09 \pm 0.73$ cm positional error in the real world. Furthermore, in the tasks of cube picking, cube stacking, and cube placing tasks, they yielded 19, 18, and 15 successful attempts out of 20.

\cite{zhang_adversarial_2019}. presented how to efficiently transfer visuo-motor policies from simulation to real-world. In their case study, a velocity-controlled 7 DoF robot arm needed to reach a blue cuboid object in a table-top scenario. They achieved a 97.8\% success rate and 1.8 cm control accuracy.

~\cite{zhang_roi-based_2019} introduced a two-stage ROI-based robotic grasp detection model focusing object overlapping scenes. They yielded 92.5\% and 83.8\% success rate, respectively in single-object and multi-object scenes. Nevertheless, using real images, they did not focus on sim2real knowledge transfer.

It is challenging to compare the works above as many aspects of the problem are different. However, in general, 80\% success rate is considered a good performance. In our case, industrial usability is an important factor, thus our aim is to reach close to 100\% success rate keeping universality as much as possible.

It is important to mention that, according to our best knowledge, even though there are existing industrial solutions for some types of robotic grasping, they cannot perform the task described in Section~\ref{sec:problem_statement}. In general, these tools either detect a tag on a palette and then move to predefined positions on the palette, or they are only capable of detecting one class of objects, or they use many real-world images. For the aforementioned reasons, the comparison of such solutions is not feasible.

\section{Approach}

In our approach, the problem is divided into two stages. In the first stage, the different objects are localized (bounding box information with classification). In the second stage, the orientations of the detected objects are estimated with convolutional neural networks trained on class-specific examples. As the plane coordinates and the 3D models of the objects are known, with the center points and the orientations, the grasping position can be calculated effortlessly. The illustration of the proposed approach is shown in Fig.~\ref{fig_cover}.

In Section~\ref{sec:approach_object_detection} the object detection model is presented, while in Section~\ref{sec:approach_orientation_estimation} the orientation estimation is described. These two stages are the main building blocks of the MOGPE RT model. In Section~\ref{sec:roi_cropping}, the region of interest (ROI) cropping algorithm is presented which connects the two stages of the model. In Section~\ref{sec:approach_pattern_matching}, the MOGPE HP model is described, which is an extension of the MOGPE RT model. Our implementation is available at\footnote{https://git.sztaki.hu/emi/grasping-pose-estimation}.

\subsection{Object Detection (Stage 1)}\label{sec:approach_object_detection}

The object detection model needs to locate and classify all the objects on an image. In robotics, the object detection model not only needs to be precise (high value of mAP, mean area under the curve) but also needs to work fast (high FPS, frame per second). The object detection stage is built on our previous work~\citep{horvath_object_2022}. For the convolutional neural network (CNN), YOLOv4~\citep{bochkovskiyYOLOv4OptimalSpeed2020} was chosen as it has an optimal accuracy-speed trade-off.

 The network is trained on domain randomized synthetic images combined with one real example\footnote{Data augmentation was applied according to~\cite{bochkovskiyYOLOv4OptimalSpeed2020}}. Instead of sequentially training the model on the source domain (synthetic images) and then fine-tuning it on the target domain (real images), the model was trained in parallel. The real data was multiplied to have equal weight in the training process to the generated synthetic data. Thus, the model learns the domain shift and the generalization simultaneously
 
 For the synthetic data generation, the 3D models of the objects are loaded into the simulator with randomized positions and randomized textures. The camera renders images from randomized positions and the labels are generated automatically, knowing the position of the objects and the camera. Furthermore, a post-processing method is executed to introduce additional artificial noise. For further details of the sim2real object detection framework, and the synthetic data generation process, the reader is referred to~\cite{horvath_object_2022}.  
 
With this method, the reality gap could be shrunk to a satisfactory level, meaning that the model is capable of accurately locating and classifying the different objects not only in simulation but in the real world as well. The data generation lasts around 0.25 - 0.5s per image, while the training takes 12h on a GeForce RTX 2080 Ti GPU. The model prediction time is above 20 FPS on a GeForce RTX 3060 GPU. 

\subsection{ROI Cropping}\label{sec:roi_cropping}

\begin{figure}[hbt!]
\centering
\includegraphics[width=3.25in]{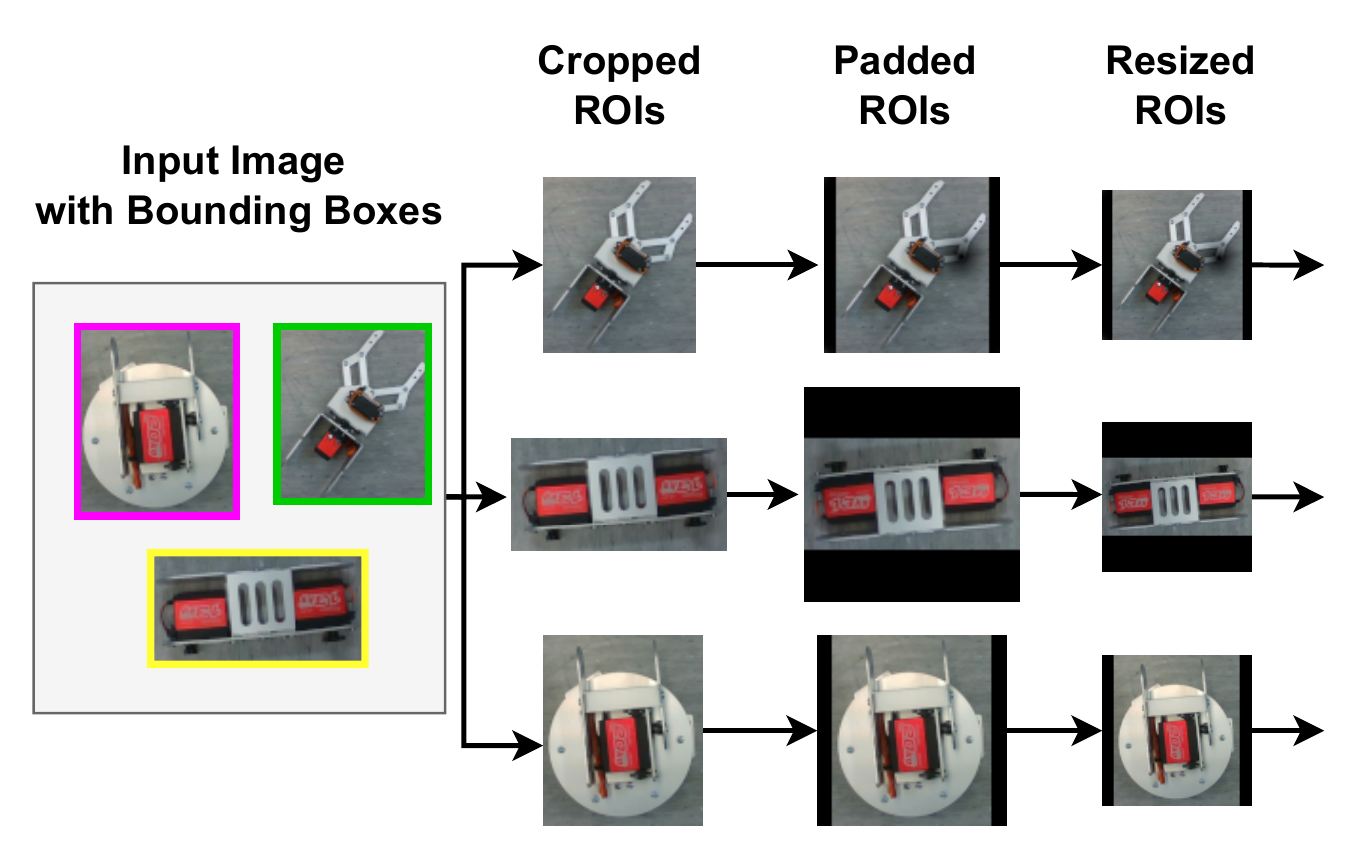}
\caption{The data flow of the ROI cropping method}
\label{fig_cropping}
\end{figure}

Between the first and second stages, a rule-based algorithm cuts out the specific ROIs of the objects from the input image according to the bounding box information. Then, it transforms them to the appropriate size while keeping the orientation of the objects (one object per image) and forwards them to the specific CNN of the second stage, depicted in Fig.~\ref{fig_cover} and detailed in Fig.~\ref{fig_cropping}. Assuming that there are $N$ classes, an object that is detected on the image can be sent to $N$ different CNNs.

As the next stage must estimate the orientation of the objects, it is crucial that the image transformation does not change the orientation. For this reason, the image is padded with zeros to a square and then resized to the expected input size of the neural network. In our case, it is 300x300.

\subsection{Orientation Estimation  (Stage 2)}\label{sec:approach_orientation_estimation}

The second stage of the model is the orientation estimation which contains $N$ CNNs (each for one class). Each of them takes a 300x300x3 image as input and outputs the sine and cosine representations of the orientation. Learning the sine and cosine values instead of learning purely the angles was chosen as the former method is a better fit for regression problems as in these trigonometric functions the distances between angles next to each other are continuous. Having computed these values, the orientation can be calculated using the \texttt{atan2} function.

\begin{figure}[hbt!]
\centering
\includegraphics[angle=90,width=2.5in]{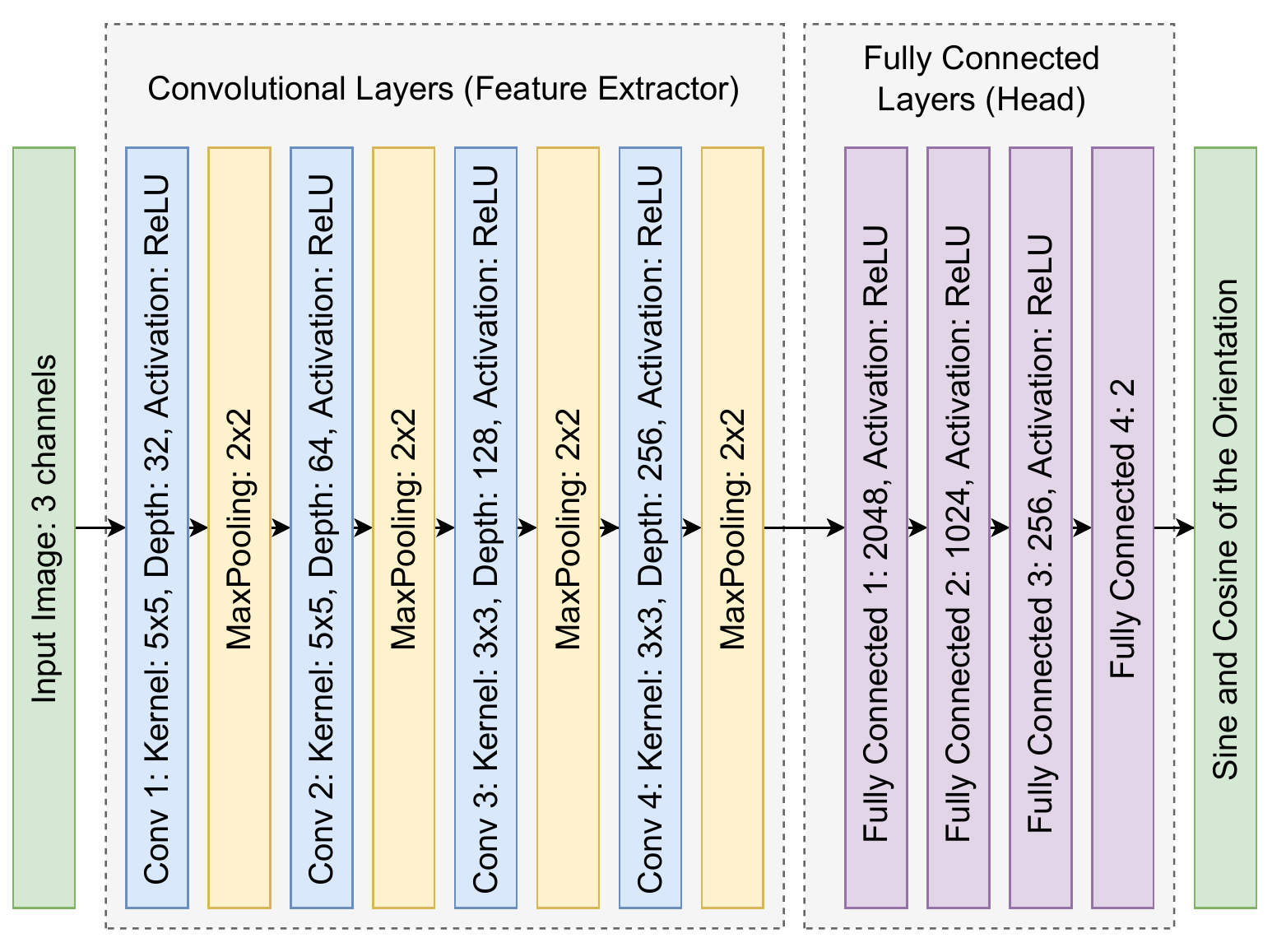}
\caption{The proposed CNN architecture for orientation estimation}
\label{fig_cnn}
\end{figure}

The architecture of the CNNs is shown in Fig.\ref{fig_cnn}. In the feature extractor, there are 4 convolutional layers with ReLU (rectified linear unit) activation functions and each of them is followed by a MaxPooling layer. To compute the outputs, there are 4 fully connected layers in the head of the network. The models are trained from scratch, independently from each other, on class-specific synthetic and real examples.

\begin{figure}[hbt!]
\centering
\includegraphics[width=2.5in]{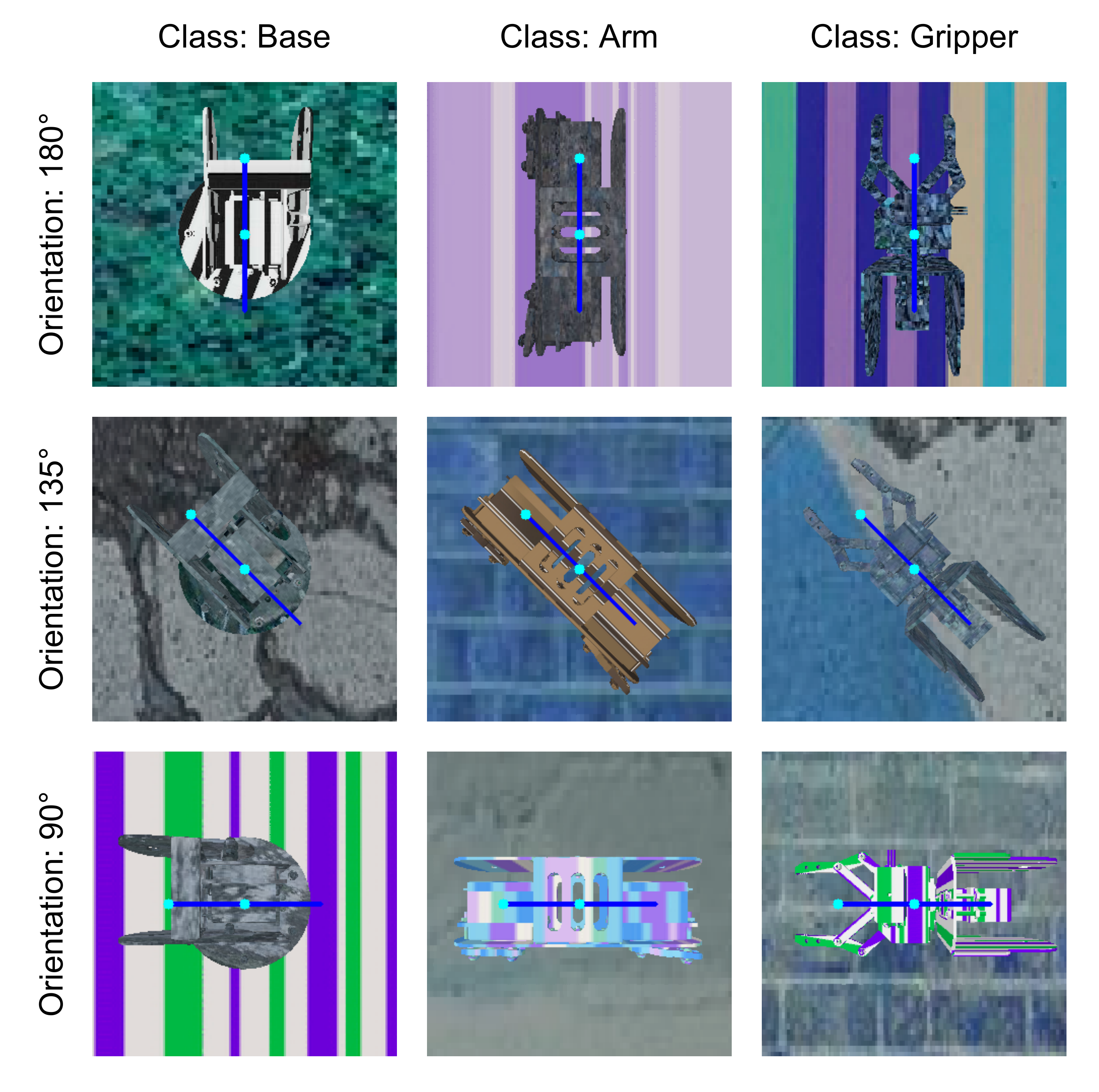}
\caption{Some examples of the generated synthetic training dataset}
\label{fig_synthetic_data}
\end{figure}

The synthetic data were generated in PyBullet. The 3D model of the object is placed in the simulator and rotated around the z-axis (perpendicular to the plane where the object is placed) while random textures are added to the plane and to the object as well. For each bit of rotation, an image is taken and the label is automatically generated with it. Some examples can be seen in Fig.~\ref{fig_synthetic_data}. The data generation lasts around 0.25 - 0.5s per image, while the training, implemented in PyTorch, takes 2.5h per class on a GeForce RTX 2080 Ti GPU. The model prediction time is at 100 FPS on a GeForce RTX 3060 GPU.

\subsection{Pattern Matching}\label{sec:approach_pattern_matching}

In the industry, it is essential that the robot grasps an object successfully and with precision. For this reason, a rule-based pattern matching algorithm is executed after the orientation estimation. The pattern matching is performed locally, in the neighborhood of the estimated orientation. With this addition, the model achieves higher precision at the cost of the extra computation. The pattern matching algorithm is the difference between the two proposed models, only the MOGPE High-Precision model incorporates this step.

The pattern matching algorithm compares the image of the object with a set of precomputed rotated kernel images. For one class, one real kernel image is rotated 359 times making 360 rotated kernel images\footnote{If the precision needs to be higher than 1 degree, this procedure can be done on a finer scale.}.

Comparing two images takes around 13 ms, thus if the search is restricted for $\pm$ 10 degrees with a 1-degree resolution, it takes 0.26 seconds. While performing it in the whole range (without the orientation estimation by the CNN) takes 4.68 seconds.

It is important to note, that the pattern matching algorithm needs a good initialization, provided by the orientation estimation CNN. Otherwise, it frequently finds wrong orientations, especially in symmetric objects.

\section{Robot Control Architecture}\label{sec:results_ros}

In this section, the robot control architecture is presented which shows how our computer vision models can be utilized in real-world robotic applications.

\begin{figure}[hbt!]
	\centering
	\includegraphics[width=3.25in]{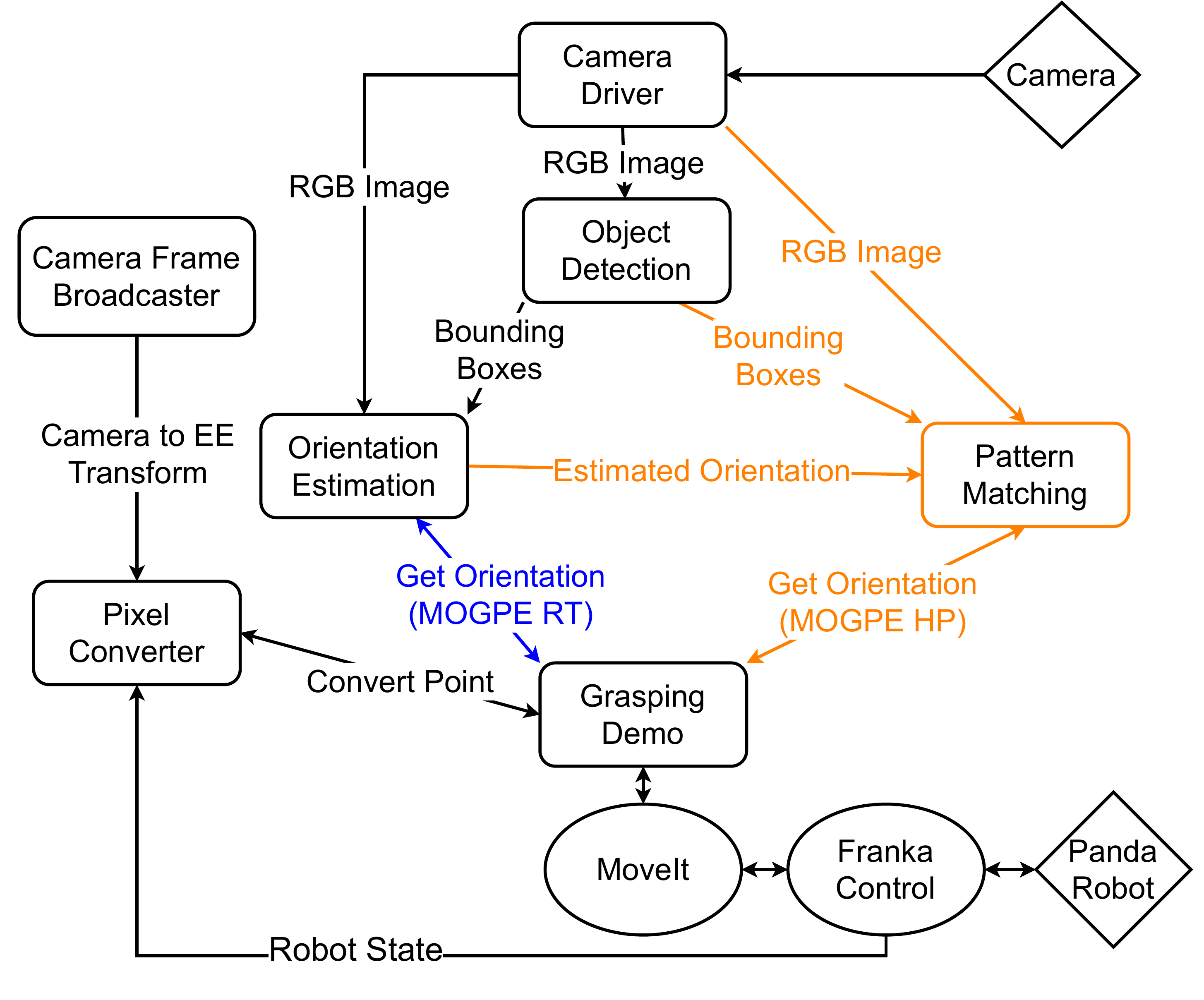}
	\caption{The robot control architecture. With blue color, the version of MOGPE RT model, while with orange color, the version of the MOGPE HP model.}
	\label{fig_ros}
\end{figure}

The robot control architecture is based on ROS (robot operating system) and is depicted in Fig.\ref{fig_ros}. The \texttt{camera driver} node publishes the images that are first read by the \texttt{object detection} node which then publishes the bounding box information. Based on these, the \texttt{orientation estimation} node predicts the orientation of the visible objects, and sends this information to the \texttt{pattern matching} node which returns with the corrected orientation estimate when the \texttt{get} \texttt{orientation} \texttt{service} is called. In case of the MOGPE RT model, the \texttt{get} \texttt{orientation} \texttt{service} returns with the original value of the \texttt{orientation estimation} node. The \texttt{camera} \texttt{frame} \texttt{broadcaster} node publishes the transformation between the camera frame and the end effector. With this information, the \texttt{pixel} \texttt{converter} node transforms pixel coordinates to the word frame when the \texttt{convert} \texttt{point} \texttt{service} is called. For motion planning, the \texttt{MoveIt} framework~\citep{gorner_moveit_2019} was used. Our implementation is available at\footnote{https://git.sztaki.hu/emi/robot\_control\_framework}

The camera is calibrated using the VISP library~\citep{marchand_visp_2005}. By taking some pictures of a known pattern (a chessboard in our case), the transformation between the robot's end effector frame and the camera frame is calculated\footnote{https://visp-doc.inria.fr/doxygen/visp-daily/tutorial-calibration-extrinsic.html}. Since the position of the plane where the objects are placed and the 3D models of the objects are known, the inverse perspective projection equations can be used to transform object positions from the image frame to the camera frame, then transform them to the world frame using the transformation matrix obtained from the calibration.

\section{Results}\label{sec:results}

In this section, the evaluate of our approach is presented. First, the settings of the experiment are described in Section~\ref{sec:results_settings}. Then, the results of the object detection (Section~\ref{sec:results_detection}), the orientation estimation models (Section~\ref{sec:results_orientation}), and the real-world robotic grasping (Section~\ref{sec:results_grasping}) are presented. 

\subsection{Setting of the Robotic Experiments}\label{sec:results_settings}

\begin{figure}[hbt!]
\centering
\includegraphics[width=2.5in]{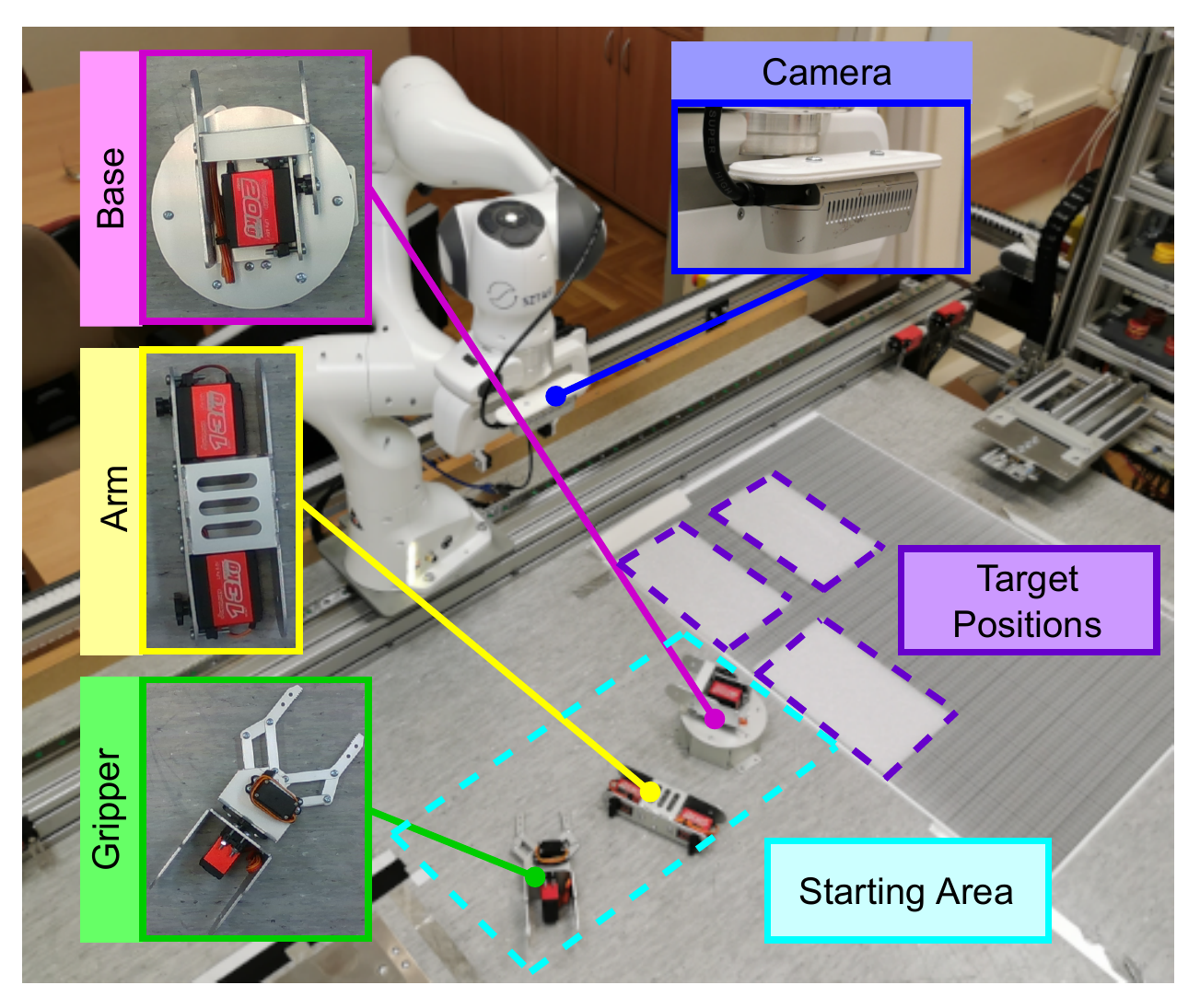}
\caption{Experimental setup.}
\label{fig_robotic_settings}
\end{figure}

For this robotic case study, three industrial parts were selected that are themselves parts of a simple robot arm, shown in Fig.~\ref{fig_robotic_settings}. Synthetic samples of the parts are depicted in Fig.~\ref{fig_synthetic_data}.

Initially, the parts are randomly placed in the starting area. The task of the robot is to pick and place the parts one by one from the starting area to the designated target positions using its two-finger gripper. Neither special illumination was applied nor monochromatic background.

\subsection{Object Detection}\label{sec:results_detection}

To train the model\footnote{Pre-trained on ImageNet}, 2000 synthetic images were generated alongside one real image (with all the 3 objects) multiplied 2000 times. The batch size was set to 64, and the other hyper-parameters of the training were chosen according to the recommendation of \cite{bochkovskiyYOLOv4OptimalSpeed2020}.

The quantitative evaluation is shown in Tab.~\ref{tab:yolo_quantitative}. The validation dataset was generated from the same distribution as the training dataset. On the other hand, the test dataset contains 59 real images, taken in different environmental and illumination conditions. Achieving 98.78\% mAP\textsubscript{50} on average can be considered a robust performance as it is close to the performance achieved on the training (99.41\% mAP\textsubscript{50}) and on the validation datasets (99.01\% mAP\textsubscript{50}). Having a reliable output of the object detection stage is crucial as this output is the input of the orientation estimation. As it is shown in our previous work ~\citep{horvath_object_2022}, the object detection part works for more classes as well, and as in the orientation estimation stage, every class processed separately, our method can be easily scaled up to more classes.

For qualitative evaluation, Fig.~\ref{fig_exm_acc_bad} shows two accurate examples of object detection. More qualitative evaluation can be found at\footnote{\label{youtube}https://youtu.be/luwA6RDEaoA}.

\begin{table}
	\begin{center}
		\caption{The mAP\textsubscript{50} scores of the object detection model}
		\begin{tabular}{p{1.7cm}|p{1.3cm}|p{1.3cm}|p{1.3cm}}
			\hline
			 &  \multicolumn{3}{c}{Dataset} \\
			& Train & Valid & Test\\
			\hline
			Training \#1 & 100\%   & 100\%    & 98.85\% \\
	    	Training \#2 & 100\%   & 100\%    & 98.81\% \\
	    	Training \#3 & 100\%   & 99.81\%  & 98.85\% \\
	    	Training \#4 & 100\%   & 100\%    & 98.81\% \\
	    	Training \#5 & 97.07\% & 95.26\%  & 98.56\% \\
			\hline
			\hline
			\textbf{AVG.} & \textbf{99.41\%}  & \textbf{99.01\%} & \textbf{98.78\%} \\
            
            \textbf{STD.} & \textbf{1.3103\%}  & \textbf{2.1001\%} & \textbf{0.1224\%} \\
			\hline
		\end{tabular}
		\label{tab:yolo_quantitative}
	\end{center}
\end{table}

\subsection{Orientation Estimation}\label{sec:results_orientation}

To train the orientation estimation CNNs, 4320 synthetic annotated images were generated. As a real dataset 15, 12, and 12 real images were available from the classes of the base, arm, and gripper. These real images were also augmented by rotating them 359 times which resulted in (with the original one) 5400, 4320, and 4320 images per class. 720 (2 times 360) real images were taken away per class for validation and testing. The loss function is MSE with Adam optimizer and the learning rate is 0.001. The batch size is 128, the training time is 100 epoch with early stopping. The loss function converged rapidly both on the training and on the validation set.

\begin{table}
	\begin{center}
		\caption{The success rate of the pose estimation model. A successful estimation is defined as within 10 degrees to the ground truth. 'Train S.' and 'Train R.' are abbreviations for the training datasets of synthetic and real images.}
		\begin{tabular}{p{1.1cm}|p{0.95cm}|p{0.95cm}|p{0.95cm}||p{0.95cm}|p{0.95cm}}
			\hline
			Dataset &  Base & Arm & Gripper & \textbf{AVG.} & \textbf{STD.} \\
			\hline
			Train S. & 99.76\% & 98.71\% & 99.54\% & \textbf{99.34\%} & \textbf{0.55\%} \\
			Train R.  & 99.85\% & 99.75\% & 99.83\% & \textbf{99.81\%}  & \textbf{0.05\%} \\
			Valid  & 100.0\% & 99.16\% & 99.72\% & \textbf{99.63\%}  & \textbf{0.42\%} \\
			Test  & 99.17\% & 92.22\% & 99.72\% & \textbf{97.04\%}  & \textbf{4.18\%} \\
			\hline
		\end{tabular}
		\label{tab:orientation_quantitative_success_rate}
	\end{center}
\end{table}

For quantitative evaluation, Tab.~\ref{tab:orientation_quantitative_success_rate} shows the success rate of the models. In the case of the base and gripper objects, the success rate is above 99\% in all datasets. In the case of the arm, the model achieves a 92\% success rate on the test dataset. The main reason behind this phenomenon is the fact that the arm object is more symmetric than the other objects. Thus, shrinking the range of estimation to 180 degrees would increase the performance. 

\begin{figure}
\centering     
\subfigure[Accurate example]{\includegraphics[width=41mm]{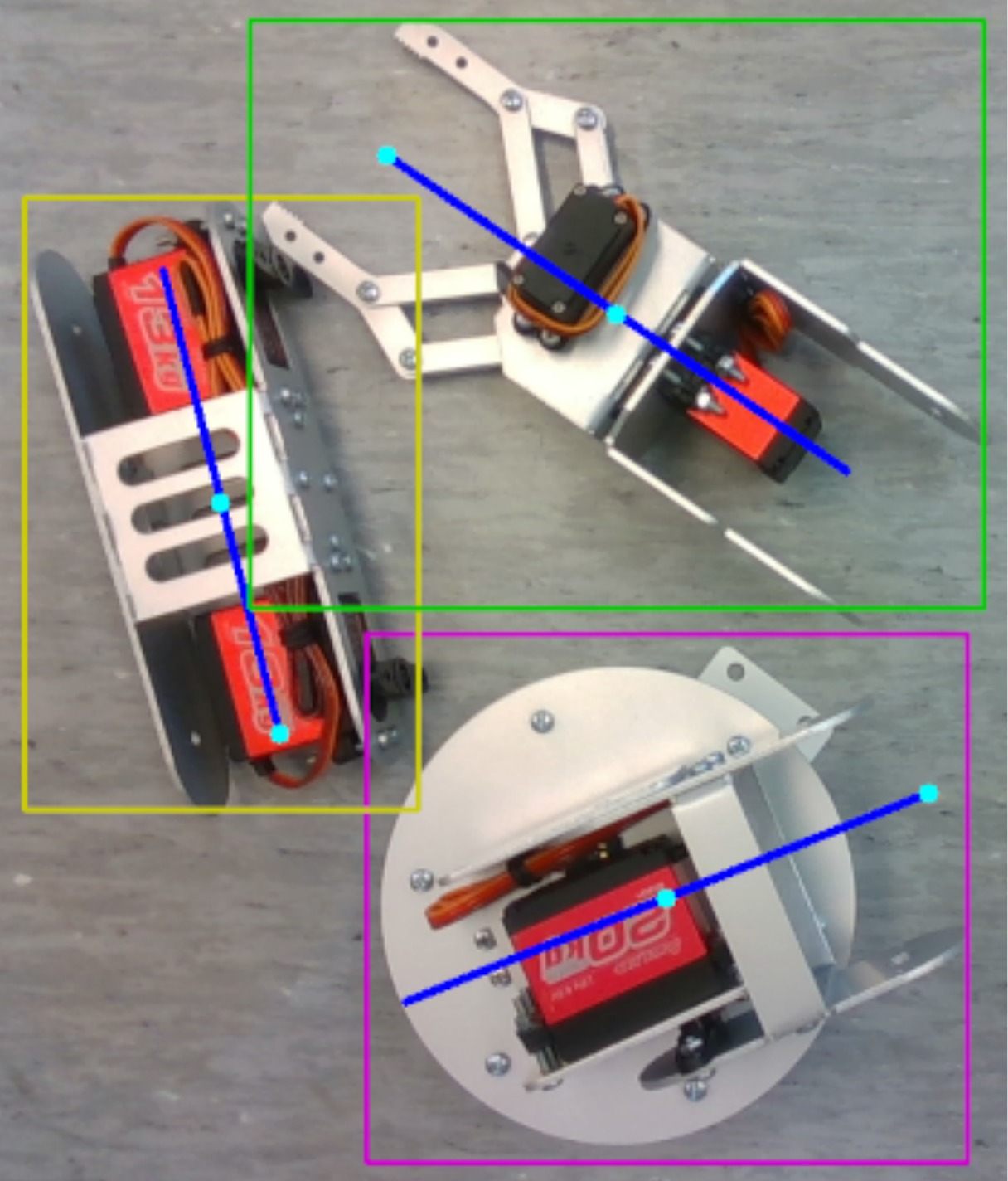}}
\subfigure[Inaccurate example]{\includegraphics[width=41mm]{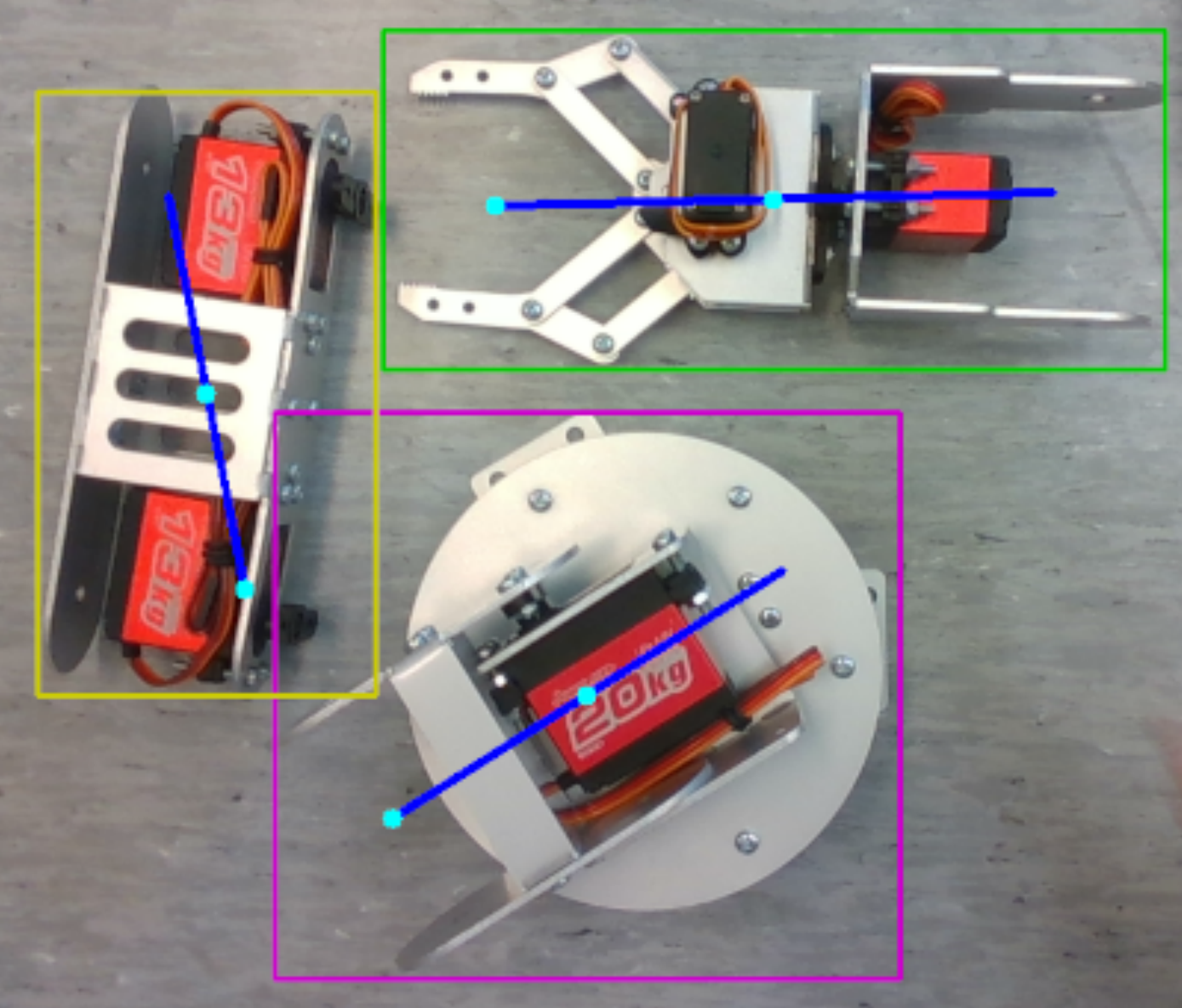}}
\caption{An accurate (a) and an inaccurate (b) prediction. The orientation of the arm is slightly tilted in the latter case. Regarding the object detection, both examples are accurate.}
\label{fig_exm_acc_bad}
\end{figure}

For qualitative evaluation, Fig.~\ref{fig_exm_acc_bad} shows an accurate and an inaccurate example of orientation estimation. More qualitative evaluation can be found at\footnote{See footnote \ref{youtube}.}.

\subsection{Robotic Grasping}\label{sec:results_grasping}

Finally, the performance of our models was measured in a real-world robotic grasping experiment using a 7 DoF collaborative robot. Ten grasp attempts were made per class and per model (all in all 60 grasp attempts). The results of the experiments are summarized in Tab.~\ref{tab:grasp_results}. The MOGPE Real-Time model worked well in the case of the arm and gripper classes. Nevertheless, it failed to reliably grasp the base class. On the other hand, the MOGPE High-Precision model could successfully grasp the objects most of the times, yielding a 96.67 \% success rate. Six grasp attempts are shown at\footnote{See footnote \ref{youtube}.}.


\begin{table}[h]
	\begin{center}
		\caption{Results of the robotic grasping experiment}
		\begin{tabular}{p{1.8cm}|p{1.0cm}|p{1.0cm}|p{1.0cm}|p{1.0cm}}
			\hline
			Model &  Base & Arm & Gripper & \textbf{Success rate} \\
			\hline
			MOGPE RT & 5/10 & 9/10 & 10/10 & \textbf{80\%} \\
			MOGPE HP & 10/10 & 10/10 & 9/10 & \textbf{96.67\%} \\
			\hline
		\end{tabular}
		\label{tab:grasp_results}
	\end{center}
\end{table}

\section{Conclusions and
future work}

In this paper, robotic grasping was addressed, a critical challenge of adaptive robotics which plays an essential role in achieving truly co-creative cyber physical systems. Two vision-based, multi-object grasp pose estimation models were presented, the MOGPE Real-Time and the MOGPE High-Precision. Furthermore, a sim2real knowledge transfer method based on domain randomization to diminish the reality gap and to overcome the data shortage. 

Our framework provides an industrial tool for fast data generation and model training and requires minimal domain-specific data. In test time, the model does not only work fast (object detection 20 FPS, orientation estimation 100 FPS) but performs well (98.78\% mAP\textsubscript{50} score, and 97.04\% success rate).

Our approach is validated not only on images but in a real-world robotic grasping experiment where the MOGPE RT model achieved an 80\%, while the MOGPE HP model accomplished a 96.67\% success rate. 

In the future, our target is to further improve our sim2real transfer learning methods expecting to gain performance with a more versatile synthetic dataset. Additionally, an interesting continuation would be to experiment with adversarial training and other industrial setups.

\begin{ack}
This research has been supported by the ED\_18-2-2018-0006 grant on an ``Research on prime exploitation of the potential provided by the industrial digitalisation''

The research was supported by the European Union within the framework of the National Laboratory for Autonomous Systems. (RRF-2.3.1-21-2022-00002)
\end{ack}

\bibliography{hdlib}             
\end{document}